\let\accentvec\vec
\documentclass[runningheads]{llncs}

\let\vec\accentvec
\usepackage[T1]{fontenc}
\usepackage{color}
\usepackage{graphicx}
\usepackage{amssymb}
\usepackage{textcomp}
\usepackage{hyperref}
\usepackage{mathtools}
\usepackage{esvect}

\newcommand{\set}[1]{\mathcal{#1}}
\begin{document}
\title{ML-CrAIST: Multi-scale Low-high Frequency Information-based Cross black Attention with Image Super-resolving Transformer}
%CrAIST: Cross-domain Attention with Image Super-resolving Transformer
\author{Alik Pramanick\orcidID{0000-0002-5993-3185} \and
Utsav Bheda\orcidID{0009-0008-7636-2610} \and
Arijit Sur\orcidID{0000-0002-9038-8138}}
\authorrunning{A. Pramanick et al.}
\institute{Indian Institute of Technology Guwahati, India\\
\email{\{p.alik,u.bheda,arijit\}}@iitg.ac.in}
\maketitle              % typeset the header of the contribution
\begin{abstract}
Recently, transformers have captured significant interest in the area of single-image super-resolution tasks, demonstrating substantial gains in performance. Current models heavily depend on the network's extensive ability to extract high-level semantic details from images while overlooking the effective utilization of multi-scale image details and intermediate information within the network. Furthermore, it has been observed that high-frequency areas in images present significant complexity for super-resolution compared to low-frequency areas. This work proposes a transformer-based super-resolution architecture called ML-CrAIST that addresses this gap by utilizing low-high frequency information in multiple scales. Unlike most of the previous work (either spatial or channel), we operate spatial and channel self-attention, which concurrently model pixel interaction from both spatial and channel dimensions, exploiting the inherent correlations across spatial and channel axis. Further, we devise a cross-attention block for super-resolution, which explores the correlations between low and high-frequency information. Quantitative and qualitative assessments indicate that our proposed ML-CrAIST surpasses state-of-the-art super-resolution methods (e.g., 0.15 dB gain @Manga109 $\times$4). Code is available on \url{https://github.com/Alik033/ML-CrAIST}.

\end{abstract}
\section{Introduction}
The task of single image super-resolution (SR)~\cite{choi2023n} remains an enduring low-level challenge that centers on the restoration of high-resolution (HR) images from degraded low-resolution (LR) inputs. As an issue with inherent ambiguity and numerous possible solutions for a given LR image, several methods have emerged in recent years to address and overcome this challenge. Numerous methods in this context use convolution neural networks (CNNs)~\cite{dai2019second,dong2015image,kim2016accurate,pramanick2024attention,tai2017memnet,zhang2018image} to improve performance in a variety of applications. These methods mostly use residual learning~\cite{kim2016accurate}, dense connections~\cite{tai2017memnet}, or channel attention~\cite{zhang2018image} to build network architectures, significantly contributing to developing super-resolution models. However, the CNN-based approach exhibits a limited receptive field due to the localized nature of convolution, which hampers the global dependencies, consequently restricting the overall performance of the model.

In recent times, the Transformer architecture, initially introduced in natural language processing (NLP), has demonstrated significant success across a wide range of high-level vision tasks~\cite{cao2022swin,carion2020end,wu2020visual}. This success is attributed to its incorporation of a self-attention mechanism, which effectively establishes global dependencies. A notable advancement in SR is SwinIR~\cite{liang2021swinir}, which presents the Swin Transformer, leading to significant enhancements over state-of-the-art CNN-based models across different standard datasets. Subsequent developments, including Swin-FIR~\cite{zhang2022swinfir}, ELAN~\cite{zhang2022efficient}, and HAT~\cite{chen2023activating}, have extended the capabilities of SwinIR by utilizing Transformers to develop various network architectures for SR tasks. These methods demonstrate that appropriately enlarging the windows for the shifted window self-attention in SwinIR can lead to obvious improvements in performance. However, the increase in computational burden becomes a significant concern as the window size grows more prominent. Furthermore, Transformer-based methods rely on self-attention and need networks with more channels than previous CNN-based methods~\cite{ahn2018fast,hui2019lightweight,kim2016deeply}.  Also, they use uni-dimensional aggregation operations (either spatial or channel) and homogeneous aggregation schemes (simple hierarchical stacking of convolution and self-attention). Wang et al.~\cite{wang2023omni} consider the above problem and design OmniSR to achieve superior performance. Despite substantial progress in super-resolution methods, they even encounter visual artifacts in the resulting images, such as inadequate texture representation and loss of details. Further, it has been observed that super-resolving high-frequency image areas are more challenging than low-frequency areas. Numerous existing SR methods work solely within the spatial domain, concentrating on improving the resolution of low-resolution pixels to obtain a high-resolution image. They often overlook the potential benefits of the frequency domain, which could offer a better method for retrieving lost high-frequency information. Also, it needs to include more texture patterns of multi-scales, which is required in SR tasks. Similar textures with multiple scales may exist within a single image at different positions. For instance, repetitive patterns at different scales (such as facades, windows, etc., in a building) may appear in various locations within a single image. The multi-scale aware framework is required to use the beneficial non-local detail, which aggregates the information from all the different scales of the LR image.

To address the above mentioned issues and achieve higher performance, this work proposes a novel super-resolution model that simultaneously exploits frequency and spatial domain information at different scales. 2D Discrete Wavelet Transformation (2dDWT) is used to analyze both the high (LH, HL, and HH) and low (LL) frequency wavelet sub-bands.  To carefully design a cross-attention block, we fuse low and high frequency information to boost SR performance. We explore the features in multiple scales and systematically combine information across all scales at each resolution level, facilitating meaningful information exchange. Simultaneously, another fusion technique is proposed to combine the high-frequency sub-bands while maintaining their unique complementary characteristics that differ from simple concatenation or averaging of the sub-bands. The major contributions of this paper are as follows:
\begin{enumerate}
    \item A novel multi-scale model is proposed by utilizing both spatial and frequency domain features that is capable to enhance the spatial resolution of an low-resolution image.
    \item In addition, a low-high frequency interaction block (LHFIB) is introduced to exchange the information between low and high frequency sub-bands through the proposed cross attention block (CAB).  
    \item A non-linear approach is proposed to fuse high-frequency sub-bands using an attention mechanism for more precise restoration of high-frequency details.
    \item Informative features are obtained from different scales using CAB while preserving the high-resolution features to represent spatial details accurately.
\end{enumerate}
\section{Related Work}
\begin{itemize}
    \item[$\blacksquare$] \textbf{Conventional CNNs for SR.} CNNs have achieved remarkable success in the task of image super-resolution. SRCNN~\cite{dong2015image} is notable as the pioneering CNN-based super-resolution method, outperforming the performance of traditional approaches (e.g., bicubic, nearest-neighbor, and bilinear interpolation). After this initial advancement, significant attention has been directed towards expanding the network depth and incorporating residual learning techniques to enhance super-resolution performance~\cite{kim2016accurate,tai2017memnet,zhang2018image}. EDSR~\cite{lim2017enhanced} further improves peak signal-to-noise ratio (PSNR) results significantly by removing the unnecessary Batch Normalization layers. Additionally, RCAN~\cite{zhang2018image} integrates a channel attention mechanism to enhance feature aggregation efficiency, enabling improved performance even with deeper network architectures. Subsequent models such as SAN~\cite{dai2019second}, NLSA~\cite{mei2021image}, and HAN~\cite{niu2020single} have introduced a range of attention mechanisms, either focusing on spatial or channel dimensions, reflecting a growing trend in attention-based approaches within the field. To improve reconstruction quality while working within constrained computing resources, DRCN~\cite{kim2016deeply}, DRRN~\cite{tai2017image}, CARN~\cite{ahn2018fast}, IMDN~\cite{hui2019lightweight} delve into lightweight architectural designs. Another research direction is operating model compression strategies like knowledge distillation~\cite{gao2018image,zhang2021data} and neural architecture search~\cite{chu2021fast} to decrease computing costs.

    \item[$\blacksquare$] \textbf{Generative adversarial networks (GANs) for SR.} GANs~\cite{goodfellow2014generative} provide a fundamental method to balance perception and distortion by regulating the weights of perceptual and fidelity losses, generating realistic images. \cite{ledig2017photo} introduced SRGAN, which incorporates adversarial training with the SRResNet generator. \cite{wang2018esrgan} presented ESRGAN featuring the residual-in-residual dense block framework for super-resolution. Later, \cite{rakotonirina2020esrgan+} enhanced ESRGAN by auxiliary noise injection and proposed ESRGAN+. Park et al.~\cite{park2023perception} suggested Flexible Style SR, which optimizes the SR model with image specific objectives without viewing the regional features. These methods~\cite{ledig2017photo,park2022flexible,rakotonirina2020esrgan+,wang2018esrgan} suffer from the computational burden posed by numerous image maps.

    \item[$\blacksquare$] \textbf{Transformer-based methods for SR.} Recently, Transformers have shown significant promise in a range of vision tasks, including image classification~\cite{wu2020visual}, object detection~\cite{carion2020end}, semantic segmentation~\cite{cao2022swin}, image restoration~\cite{chen2021pre,liu2022learning,wang2022uformer}, etc. Among these approaches, the most prominent example is the Vision Transformer (ViT), demonstrating that transformers can outperform convolutional neural networks in feature encoding tasks. Designing transformer-based models for image super-resolution poses a significant challenge as it requires preserving the structural details of the input image. IPT~\cite{chen2021pre} is a pre-trained model built upon the transformer encoder and decoder structure and has been used for super-resolution. SwinIR~\cite{liang2021swinir} employs a window-based attention mechanism to tackle image super-resolution tasks, demonstrating superior performance over CNN-based methods. ELAN~\cite{zhang2022efficient} facilitates the architecture of SwinIR and utilizes self-attention calculated in different window sizes to capture correlations between long-range pixels. Choi et al.~\cite{choi2023n} introduce N-gram context into low-level vision tasks using Transformers for the SR task. Most recently, OmniSR~\cite{wang2023omni} explored spatial-channel axis aggregation networks to enhance SR performance.
\end{itemize}
Our approach also relies on the transformer architecture. Unlike the aforementioned methods, which predominantly utilize spatial domain information and compute self-attention for model construction, our primary focus is on leveraging spatial-frequency domain features and multi-scale features through cross-attention to improve the performance of the super-resolution model.  
\begin{figure*}[ht]
    \centering
    \setlength\abovecaptionskip{-0.1\baselineskip}
    \includegraphics[width=\textwidth]{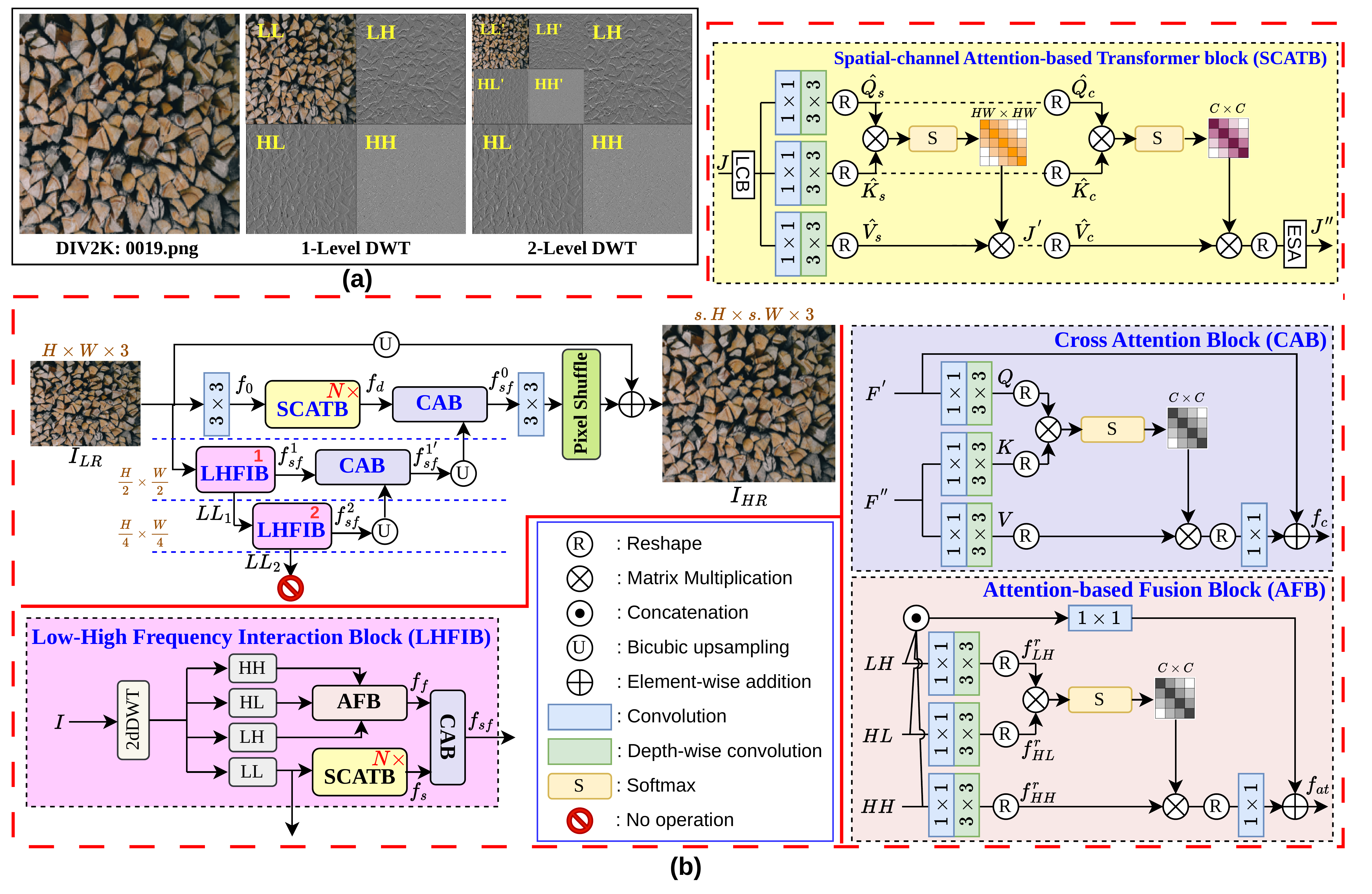}
    \caption{(a) Multi-level wavelet sub-bands of a LR image. (b) Overview of the Proposed ML-CrAIST. \textcolor{red}{$N\times$} indicates that the block is stacked N times.}
    \label{fig:craist}
\end{figure*}
\section{Proposed Method}
Figure \ref{fig:craist} shows the proposed architecture that aims to generate an SR image from the degraded LR image.
\subsection{Overall Pipeline}
This section presents a comprehensive description of the overall network architecture. Given an LR image $I_{LR} \in \mathbb{R}^{H \times W \times 3}$, we pass it through a convolution layer to extract the initial feature $f_0$. The acquired feature is then fed into $N$ spatial-channel attention-based transformer blocks (SCATB), from which the deep spatial and channel-wise correlated features $f_d$ are extracted.
\begin{equation}
    f_0 = \set{C}^{3\times3}(I_{LR}), \quad f_i = \set{F}^i_{SCATB}(f^{i-1}), \quad f_d = f_N
\end{equation}
where $\set{C}^{3\times 3}$ refers a convolution with $3\times 3$ kernel size, $\set{F}^i_{\set{SCATB}}$ represents the $i$-th SCATB, and $f_1,f_2,..,f_N$ denote intermediate features.

Simultaneously, we input $I_{LR}$ into the first low-high frequency interaction block (LHFIB) to extract spatial-frequency information $f^1_{sf}$ $ \in \mathbb{R}^{\frac{H}{2} \times \frac{W}{2} \times c}$ and LL cube. The LL cube of the first LHFIB is fed into the second LHFIB to extract further spatial-frequency information ($f^2_{sf}$ $\in \mathbb{R}^{\frac{H}{4} \times \frac{W}{4} \times c}$) in different scales. Each LHFIB contains an attention-based fusion block (AFB) to fuse the high-frequency sub-bands, $N$ number of SCATBs to capture spatially and channel-wise refined features from the low-frequency sub-band, and a cross attention block (CAB) for message passing between refined low-high frequency features. Next, we up-sample $f^2_{sf}$ and combine it with $f^1_{sf}$ within the cross-attention block (CAB) to obtain informative multi-scale features, denoted as $f_{sf}^{1'} \in \mathbb{R}^{\frac{H}{2} \times \frac{W}{2} \times c}$. Next, we up-sample the $f_{sf}^{1'}$ feature and fed it alongside $f_d$ into the CAB module to generate meaningful features ($f^0_{sf} \in \mathbb{R}^{H \times W \times c}$)  that contain refined multi-scales feature information.
\begin{equation}
\begin{split}
    f^1_{sf}, LL_1 = \set{F}_{LHFIB}(I_{LR}), \quad  f^2_{sf}, LL_2 = \set{F}_{LHFIB}(LL_1), \\
    f_{sf}^{1'} = \set{F}_{CAB}(f^1_{sf}, \set{U}(f^2_{sf})), \quad f^0_{sf} = \set{F}_{CAB}(f_d, \set{U}(f^1_{sf}))
\end{split}
\end{equation}
where, $\set{F}_{LHFIB}$, $\set{F}_{CAB}$, and $\set{U}$ represent the LHFIB, CAB and bicubic up-sampling operation. Next, we employ a convolution layer and set the output channels to $3s^2$ , where $s$ denotes the scale factor by which the spatial resolution is to be enhanced. Finally, a PixelShuffle ($\set{PS}$) layer takes the low-resolution feature maps ($f_l \in \mathbb{R}^{H \times W \times 3s^2}$) and produce the high-resolution image $I_{HR} \in \mathbb{R}^{s.H \times s.W \times 3}$. Then, the reconstructed HR image $I_{HR}$ can be written as
\begin{equation}
    I_{HR} = \mathcal{PS}(f_l) + \mathcal{U}(I_{LR}), \quad f_l = \mathcal{C}^{3\times 3}(f^0_{sf})
\end{equation}
The proposed ML-CrAIST is optimized using the $\mathcal{L}_1$ loss:
\begin{equation}
    \mathcal{L}_1(I^g_{HR}, I_{HR}) =  \frac{1}{M}\sum_{a=1}^{M}\|(I_{HR})^a - (I^g_{HR})^{a}\|_1
\end{equation}
where $I^g_{HR}$ indicates the ground-truth image.
\subsection{Spatial-channel attention-based transformer block (SCATB)}
Wang et al.~\cite{wang2023omni} introduced the omni-self attention (OSA) block, which has been integrated to capture pixel interactions from spatial and channel dimensions simultaneously, enabling the exploration of potential correlations across spatial and channel dimensions. Instead of using a standard transformer block, we leverage the OSA block along with LCB~\cite{wang2023omni} and ESA~\cite{kong2022residual} block as a SCATB to capture useful local details and long-range dependencies effectively.

To formally define its operational principle,  let $J \in \mathbb{R}^{H \times W \times C}$ be the intermediate feature map that passes through an LCB block ($\mathcal{F}_{LCB}$) to aggregate local contextual information ($f_c^l$),  then SCATB generates query (Q), key (K) and value (V) projections by using a $1 \times 1$ convolution ($\mathcal{C}^{1\times 1}$) followed by $3 \times 3$ depth-wise convolution ($\mathcal{D}_c^{3 \times 3}$) on $f_c^l$, where $Q,K,V \in \mathbb{R}^{H \times W \times C}$. Next, we reshaped query ($\hat{Q_s} \in \mathbb{R}^{HW \times C}$), key ($\hat{K_s} \in \mathbb{R}^{C \times HW}$), and value ($\hat{V_s}\in \mathbb{R}^{HW \times C}$) projections, and calculate the attention map of size $\mathbb{R}^{HW \times HW}$ between $\hat{Q_s}$ and $\hat{K_s}$ in spatial dimension which is multiplied with the $\hat{V_s}$ to get the spatially enriched attentive features $J^{'}$. Next stage, to get the attention map of size $\mathbb{R}^{C \times C}$ in channel dimension, we reshape query ($\hat{Q_c} \in \mathbb{R}^{C \times HW}$), key ($\hat{K_c} \in \mathbb{R}^{HW \times C}$) and $J^{'}$ as value ($\hat{V_c} \in \mathbb{R}^{C \times HW}$). Then, we perform the matrix multiplication between $\hat{Q_c}$ and $\hat{K_c}$ followed by a softmax operation and get the channel-wise attentive feature map. Finally, the channel-wise attentive feature maps are multiplied with the $\hat{V_c}$ and get the spatial and channel-wise correlated feature maps. Lastly, these feature maps are fed into the ESA block ($\mathcal{F}_{ESA}$) to refine the features further. Overall, the procedure is described as: 
\begin{equation}
    \begin{split}
        Q,K,V = \mathcal{D}_c^{3\times3}(\mathcal{C}^{1\times1}(f_c^l)), \quad f_c^l = \mathcal{F}_{LCB}(J),\quad \hat{K_s} = \mathcal{R}(K),\\ \hat{Q_s}=\mathcal{R}(Q),\quad \hat{V_s}=\mathcal{R}(V),\quad J^{'} = \mathcal{S}(\hat{K_s}.\hat{Q_s}).\hat{V_s}, \quad \hat{K_c}=\mathcal{R}(\hat{K_s}),\\\hat{Q_c}=\mathcal{R}(\hat{Q_s}),\quad \hat{V_c}=\mathcal{R}(J^{'}),
        \quad J^{''} = \mathcal{F}_{SCATB}(J) = \mathcal{F}_{ESA}(\mathcal{S}(\hat{K_c}.\hat{Q_c}).\hat{V_c}),
    \end{split}
\end{equation}
where $\mathcal{S}$, $\mathcal{R}$, and $\mathcal{F}_{SCATB}$, indicate the softmax function, reshape, and spatial-channel attention-based transformer operation, respectively.
We encourage the reader to refer~\cite{wang2023omni} for more details. We have demonstrated that OSA is advantageous over standard transformer block~\cite{zamir2022restormer} in producing visually pleasing SR images in the experiments section.
\subsection{Low-High Frequency Interaction Block (LHFIB)}
In this work, to integrate frequency domain information with spatial domain, we apply the Haar wavelet transformation as a 2D discrete wavelet transformation to the LR image ($I_{LR}$) and decompose it into four sub-bands (LL, LH, HL, and HH) where every sub-band  $\in \mathbb{R}^{\frac{H}{2} \times \frac{W}{2} \times 1}$. The LL sub-band characterizes the background details within the image, while LH, HL, and HH sub-bands characterize variations along vertical axis, variations along horizontal axis, and diagonal information present in the image. The LL sub-band and the original degraded image are typically employed for analyzing spatial information. Since LH, HL, and HH sub-bands preserve high-frequency components, they provide richer content for enhancing high-frequency detail during the super-resolution process. To leverage the benefit of frequency and spatial details, we design a low-high frequency interaction block. 

In detail, let it take $I$ as input and break it down into $LL, LH, HL$, and $HH$ components. Next, we combine the high-frequency sub-bands (i.e., $LH, HL$, and $HH$) using an attention-based fusion block (AFB) and get the refined high-frequency information $f_f$. The low-frequency (i.e., LL) sub-band is fed into SCATB to extract useful spatial information $f_s$. Finally, we have performed the cross-attention between low and high-frequency features to enable intelligent feature aggregation. The entire approach can be formulated as:
\begin{equation}
    \begin{split}
        {LL}, {LH}, {HL}, {HH} = \mathcal{F}_{DWT}(I), \quad f_f = \mathcal{F}_{AFB}({LH},{HL},{HH}),\\
        f_s = \mathcal{F}_{SCATB}({LL}), \quad f_{sf} = \mathcal{F}_{CAB}(f_f, f_s),\quad
        f_{sf}, {LL} = \mathcal{F}_{LHFIB}(I)
    \end{split}
\end{equation}
where $\mathcal{F}_{DWT}$, $\mathcal{F}_{AFB}$, $\mathcal{F}_{CAB}$ and $\mathcal{F}_{LHFIB}$ refer 2dDWT, attention-based fusion, cross-attention, and low-high frequency interaction operation, respectively.
\subsection{Attention-based fusion block (AFB)}
 The conventional method for feature aggregation typically involves either simple concatenation or summation. However, these types of selection offer restricted expressive capabilities of the network, as \cite{li2019selective} suggested. In this context, we present a nonlinear method for merging features through an attention mechanism to identify and amplify the more relevant features. As shown in Figure \ref{fig:craist}, we propose an attention-based fusion block (AFB) to combine the high-frequency cubes so that only useful information can be processed further. We pass the high-frequency sub-bands through a convolution layer with $1 \times 1$ kernel size and a depth-wise convolution layer with $3 \times 3$ kernel size. Next, we reshape the features to obtain $f^r_{LL}, f^r_{HH} \in \mathbb{R}^{C \times HW}$ and $f^r_{HL} \in \mathbb{R}^{HW \times C}$. We compute the matrix multiplication between $f^r_{LH}$  and $f^r_{HL}$ followed by a softmax operation to get the attentive map ($f_a$) of size $\mathbb{R}^{C \times C}$. This attention map $f_a$ is multiplied with $f^r_{HH}$ to obtain attentive feature $f_{at}$. Finally, the concatenated LH, HL, and HH sub-bands are convolved through a $1 \times 1$ convolution and added with the reshaped attentive feature to produce the attention-based fused high-frequency features. Such an operation can be defined as: 
\begin{equation}
    \begin{split}
        f_{sb} = \mathcal{D}_c^{3 \times 3}(\mathcal{C}^{1 \times 1}(sb)),\quad f^r_{sb}=\mathcal{R}(f_{sb}), \quad sb \in {\{LH, HL, HH\}}\\
        f_{at} = \mathcal{F}_{AFB}(LH,HL,HH) = \mathcal{C}^{1\times 1}(LH \odot HL \odot HH) \\ + \mathcal{C}^{1 \times 1}(\mathcal{R}(\mathcal{S}(f^r_{LH}.f^r_{HL}).f^r_{HH})),
    \end{split}
\end{equation}
where $\mathcal{S}$, $\mathcal{R}$, $\odot$ refer to softmax function, reshape operation, and concatenation operation, respectively. Through ablation, we have shown that the AFB yields more promising outcomes than regular concatenation and addition.
\subsection{Cross Attention Block (CAB)}
CAB integrates two distinct embedding sequences of identical dimensions. It employs query from one sequence and key and value from the other. The attention masks from one embedding sequence are used to emphasize the extracted features in another embedding sequence. We introduce two cross-attention blocks (CAB) with similar architectures for message passing: one operates between low-high frequency features, and the other operates between multi-scale features. For low-high frequency features, it leverages the low frequency features ($F^{'}$) to generate a query projection and employs high frequency features ($F^{''}$) to create key and value projections through a standard $1 \times 1$ convolution and a $3 \times 3$ depth-wise convolution layer. Similarly, in the multi-scale scenario, one scale feature ($F^{'}$) is used to generate the query projection, while another scale feature ($F^{''}$) is used to create the key and value projections.  Overall, cross-attention can be obtained by
\begin{equation}
\begin{split}
    Q = \mathcal{D}_c^{3 \times 3}(\mathcal{C}^{1 \times 1}(F^{'})), \quad K,V = \mathcal{D}_c^{3 \times 3}(\mathcal{C}^{1 \times 1}(F^{''})), \quad 
    Q_r = \mathcal{R}(Q),\\ K_r = \mathcal{R}(K), \quad V_r = \mathcal{R}(V), \quad \mathcal{CA}(Q_r,K_r,V_r) = \mathcal{S}(Q_r\cdot K_r)\cdot V_r, \\
    f_c = \mathcal{F_{CAB}}(Q,K,V) = \mathcal{C}^{1 \times 1}(\mathcal{R}(\mathcal{CA}(Q_r,K_r,V_r))) + F^{'},
\end{split}
\end{equation}
where $Q, V \in \mathbb{R}^{C \times HW}$, $K \in \mathbb{R}^{HW \times C}$, and $\mathcal{CA}$ represents the cross-attention function. 
\section{Experiments}
\subsection{Datasets \& Evaluation Metrics}
Following prior research~\cite{liang2021swinir,choi2023n,wang2023omni}, we employ the DIV2K dataset \cite{timofte2017ntire} for training. For testing purposes, we utilize five widely recognized benchmark datasets: Set5 \cite{bevilacqua2012low}, Set14~\cite{zeyde2012single}, B100 \cite{martin2001database}, Urban100 \cite{huang2015single}, and Manga109 \cite{matsui2017sketch}. The testing results are assessed based on PSNR and structural similarity index measure (SSIM) values computed on the Y channel (i.e., luminance) within the YCbCr color space. Also, we evaluate the learned perceptual image patch similarity (LPIPS) metrics. It measures how similar two images appear to the human visual system.
\vspace{-15pt}
\subsection{Implementation Details}
We augment the data during training by applying random horizontal flips and 90/180/270-degree rotations. For a fair comparison with the existing works, LR images are obtained through bi-cubic down-sampling from HR images. Empirically, the number of SCATBs in ML-CrAIST is set to $5$. Also, the attention head number is set to $4$, and the window size is set to 8. We train the model using the Adam optimizer with a batch size of 32 for 1000K iterations, starting with an initial learning rate of $10^{-4}$, which is decreased by half after every 200k iterations. During each training iteration, LR patches of size $64 \times 64$ are randomly cropped as input. We have set the number of channels 64 in each convolution layer for ML-CrAIST (Ours). The proposed work is implemented using PyTorch, and all experimentations are performed on a single NVIDIA V100 GPU. Figure \ref{fig:vc2}(b) shows the convergence of the model that we observed. 

In our lighter version of ML-CrAIST (Ours-Li), we have used the same architecture shown in Figure \ref{fig:craist} with a reduced number of channels in each convolution layer from 64 to 48. 
\begin{table}
    %\scriptsize
    \tiny
    \centering
    \setlength{\tabcolsep}{1pt}
    \begin{tabular}{c|c|c|c|c|c|c|c|c|c|c|c|c|c|c}
        \hline
         & & & \#params& FLOPs&\multicolumn{2}{|c|}{\textbf{Set5}} & \multicolumn{2}{|c|}{\textbf{Set14}} & \multicolumn{2}{|c|}{\textbf{B100}} & \multicolumn{2}{|c|}{\textbf{Urban100}} & \multicolumn{2}{|c}{\textbf{Manga109}} \\
        \hline
        Method &  & Years & (K)&(G) & PSNR & SSIM & PSNR & SSIM & PSNR & SSIM & PSNR & SSIM & PSNR & SSIM  \\
        \hline
        VDSR &  & CVPR'16 & 666&613 & 36.66 & 0.9542 & 33.05 & 0.9127 & 31.90 & 0.8960 & 30.76 & 0.9140 & 37.22 & 0.9750  \\
        MemNet &  & ICCV'17 & 678& 2662.4& 37.78 & 0.9597 & 33.28 & 0.9142 & 32.08 & 0.8978 & 31.31 & 0.9195 & 37.72 & 0.9740\\
        SRMDNF &  & CVPR'18 & 1511&- & 37.79 & 0.960 & 33.32 & 0.915 & 32.05 & 0.8985 & 31.33 & 0.9204 & 38.07 & 0.9761\\
        CARN &  & ECCV'18 & 1592& 222.8&37.76 & 0.9590 & 33.52 & 0.9166 & 32.09 & 0.8978 & 31.92 & 0.9256  & 38.36 & 0.9765 \\
        IMDN & & MM'19 & 694& 158.8&38.00 & 0.9605 & 33.63 & 0.9177 & 32.19 & 0.8996 & 32.17 & 0.9283 & 38.88 & 0.9774 \\
        LatticeNet & $2\times$ & ECCV'20 & 756& 169.5& 38.06 & 0.9610 & 33.78 & 0.9193 & 32.25 & 0.9005 & 32.43 & 0.9302 & 38.94 & 0.9774\\
        SwinIR &  & ICCVW'21 & 878& 195.6& \textcolor{black}{38.14} & \textcolor{black}{0.9611} & \textcolor{blue}{33.86} & \textcolor{black}{0.9206} & \textcolor{green}{32.31} & \textcolor{green}{0.9012} & \textcolor{black}{32.76} & \textcolor{black}{0.9340} & 39.12 & \textcolor{black}{0.9783}\\
        ESRT &  & CVPRW'22 & 677& 191.4& 38.03 & 0.9600 & 33.75 & 0.9184 & 32.25 & 0.9001 & 32.58 & 0.9318 & \textcolor{black}{39.12} & 0.9774\\
        NGSwin &  & CVPR'23 & 998& 140.4& \textcolor{black}{38.05} & \textcolor{black}{0.9610} & \textcolor{green}{33.79} & \textcolor{black}{0.9199} & \textcolor{black}{32.27} & \textcolor{black}{0.9008} & 32.53 & 0.9324 & 38.97 & 0.9777 \\
        OmniSR &  & CVPR'23 & 772& 147.2&\textcolor{red}{38.22} & \textcolor{green}{0.9613} & \textcolor{red}{33.98} & \textcolor{green}{0.9210} & \textcolor{red}{32.36} & \textcolor{blue}{0.9020} & \textcolor{red}{33.05} & \textcolor{blue}{0.9363} & \textcolor{red}{39.28} & \textcolor{green}{0.9784} \\
        \hline
        \textbf{Ours-Li} & & & \textbf{743}& \textbf{97.2}&\textcolor{green}{38.15} & \textcolor{blue}{0.9615} & \textcolor{black}{33.64} & \textcolor{blue}{0.9213} & \textcolor{blue}{32.35} & \textcolor{blue}{0.9020} & \textcolor{green}{32.93} & \textcolor{green}{0.9361} & \textcolor{green}{39.23} & \textcolor{blue}{0.9785} \\
        \textbf{Ours} & & & 1259& 165.7&\textcolor{blue}{38.19} & \textcolor{red}{0.9617} & \textcolor{black}{33.77} & \textcolor{red}{0.9220} & \textcolor{red}{32.36} & \textcolor{red}{0.9022} & \textcolor{blue}{33.04} & \textcolor{red}{0.9370} & \textcolor{blue}{39.26} & \textcolor{red}{0.9786} \\
        \hline
        \hline
        VDSR &  & CVPR'16 & 666& 613&33.66 & 0.9213 & 29.77 & 0.8314 & 28.82 & 0.7976 & 27.14 & 0.8279 & 32.01 & 0.9340 \\
        MemNet &  & ICCV'17 & 678& 2662.4&34.09 & 0.9248 & 30.00 & 0.8350 & 28.96 & 0.8001 & 27.56 & 0.8376 & 32.51 & 0.9369\\
        EDSR &  & CVPRW'17 & 1555& 160.2&34.37 & 0.9270 & 30.28 & 0.8417 & 29.09 & 0.8052 & 28.15 & 0.8527 & 33.45 & 0.9439 \\
        SRMDNF &  & CVPR'18 & 1528& -&34.12 & 0.9254 & 30.04 & 0.8382 & 28.97 & 0.8025 & 27.57 & 0.8398 & 33.00 & 0.9403 \\
        CARN &  & ECCV'18 & 1592& 118.8&34.29 & 0.9255 & 30.29 & 0.8407 & 29.06 & 0.8034 & 28.06 & 0.8493 & 33.50 & 0.9440 \\
        IMDN &  & MM'19 & 703& 56.3&34.36 & 0.9270 & 30.32 & 0.8417 & 29.09 & 0.8046 & 28.17 & 0.8519 & 33.61 & 0.9445 \\
        RFDN-L & $3\times$ & ECCV'20 & 633& 65.6& 34.47 & 0.9280 &  30.35 & 0.8421 & 29.11 & 0.8053 & 28.32 & 0.8547  & 33.78 & 0.9458 \\
        LatticeNet &  & ECCV'20 & 765& 76.3&34.40 & 0.9272 & 30.32 & 0.8416 & 29.10 & 0.8049 & 28.19 & 0.8513 & 33.63 & 0.9442\\
        SwinIR &  & ICCVW'21 & 886 & 87.2& \textcolor{blue}{34.62} & \textcolor{green}{0.9289} & \textcolor{blue}{30.54} & \textcolor{black}{0.8463} & \textcolor{green}{29.20} & \textcolor{black}{0.8082} & \textcolor{black}{28.66} & \textcolor{black}{0.8624} & \textcolor{black}{33.98} & \textcolor{black}{0.9478}\\
        ESRT &  & CVPRW'22 & 770& 96.4&34.42 & 0.9268 & 30.43 & 0.8433 & 29.15 & 0.8063 & 28.46 & 0.8574 & 33.95 & 0.9455\\
        NGSwin &  & CVPR'23 & 1007& 66.6& \textcolor{black}{34.52} & \textcolor{black}{0.9282} & \textcolor{green}{30.53} & 0.8456 & 29.19 & 0.8078 & 28.52 & 0.8603 & 33.89 & 0.9470 \\
        OmniSR &  & CVPR'23 & 780& 74.4&\textcolor{red}{34.70} & \textcolor{blue}{0.9294} &  \textcolor{red}{30.57} & \textcolor{green}{0.8469} & \textcolor{blue}{29.28} & \textcolor{green}{0.8094} & \textcolor{blue}{28.84} & \textcolor{blue}{0.8656}  & \textcolor{green}{34.22} & \textcolor{green}{0.9487} \\
        \hline
        \textbf{Ours-Li} & & & \textbf{749}& \textbf{49.6}&\textcolor{green}{34.58} & \textcolor{blue}{0.9294} & \textcolor{black}{30.23} & \textcolor{blue}{0.8474} & \textcolor{blue}{29.28} & \textcolor{blue}{0.8106} & \textcolor{green}{28.73} & \textcolor{green}{0.8651} & \textcolor{blue}{34.26} & \textcolor{blue}{0.9492} \\
        \textbf{Ours} & & & 1268& 84.1&\textcolor{red}{34.70} & \textcolor{red}{0.9302} & \textcolor{black}{30.39} & \textcolor{red}{0.8488} & \textcolor{red}{29.31} & \textcolor{red}{0.8111} & \textcolor{red}{28.89} & \textcolor{red}{0.8676} & \textcolor{red}{34.42} & \textcolor{red}{0.9501} \\
        \hline
        \hline
        VDSR &  & CVPR'16 & 666& 613&31.35 & 0.8838 & 28.01 & 0.7674 & 27.29 & 0.7251 & 25.18 & 0.7524 & 28.83 & 0.8870 \\
        MemNet &  & ICCV'17 & 678& 2662.4&31.74 & 0.8893 & 28.26 & 0.7723 & 27.40 & 0.7281 & 25.50 & 0.7630 & 29.42 & 0.8942 \\
        EDSR &  & CVPRW'17 & 1518& 114.0&32.09 & 0.8938 &  28.58 & 0.7813 & 27.57 & 0.7357 & 26.04 & 0.7849 & 30.35 & 0.9067 \\
        SRMDNF &  & CVPR'18 & 1552& -&31.96 & 0.8925 & 28.35 & 0.7787 & 27.49 & 0.7337 & 25.68 & 0.7731 & 30.09 & 0.9024\\
        CARN &  & ECCV'18 & 1592& 90.9&32.13 & 0.8937 & 28.60 & 0.7806 &  27.58 & 0.7349 & 26.07 & 0.7837  & 30.47 & 0.9084 \\
        IMDN &  & MM'19 & 715& 40.9&32.21 & 0.8948 & 28.58 & 0.7811 & 27.56 & 0.7353 & 26.04 & 0.7838 & 30.45 & 0.9075 \\
        RFDN-L & $4\times$ & ECCV'20 & 643& 37.4& 32.28 & 0.8957 & \textcolor{black}{28.61} & 0.7818 & 27.58 & 0.7363 & 26.20 & 0.7883 & 30.61 & 0.9096 \\
        LatticeNet &  & ECCV'20 & 777& 43.6&32.30 & 0.8962 & 28.68 & 0.7830 & 27.62 & 0.7367 & 26.25 & 0.7873 & 30.54 & 0.9075\\
        SwinIR &  & ICCVW'21 & 897&49.6 &\textcolor{blue}{32.44} & \textcolor{green}{0.8976} & \textcolor{blue}{28.77} & \textcolor{black}{0.7858} & \textcolor{black}{27.69} & \textcolor{black}{0.7406} & \textcolor{black}{26.47} & \textcolor{black}{0.7980} & \textcolor{black}{30.92} & \textcolor{green}{0.9151}\\
        ESRT &  & CVPRW'22 & 751& 67.7&32.19 & 0.8947 & \textcolor{green}{28.69} & 0.7833 & \textcolor{black}{27.69} & 0.7379 & 26.39 & 0.7962 & 30.75 & 0.9100\\
        NGSwin &  & CVPR'23 & 1019& 36.4&\textcolor{black}{32.33} & \textcolor{black}{0.8963} & \textcolor{red}{28.78} & \textcolor{green}{0.7859} & 27.66 & 0.7396 & 26.45 & 0.7963 & 30.80 & \textcolor{black}{0.9128} \\
        OmniSR &  & CVPR'23 & 792& 37.8&\textcolor{red}{32.49} & \textcolor{red}{0.8988} & \textcolor{red}{28.78} & \textcolor{green}{0.7859} & \textcolor{green}{27.71} & \textcolor{green}{0.7415} & \textcolor{blue}{26.64} & \textcolor{green}{0.8018} & \textcolor{green}{31.02} & \textcolor{green}{0.9151} \\
        \hline
        \textbf{Ours-Li} & & & \textbf{758}& \textbf{25.5}&\textcolor{black}{32.15} & \textcolor{black}{0.8962} & \textcolor{black}{28.40} & \textcolor{blue}{0.7863} & \textcolor{blue}{27.73} & \textcolor{blue}{0.7426} & \textcolor{green}{26.53} & \textcolor{blue}{0.8019} & \textcolor{blue}{31.11} & \textcolor{blue}{0.9162} \\
        \textbf{Ours} & & & 1280& 42.9&\textcolor{green}{32.36} & \textcolor{blue}{0.8984} & \textcolor{black}{28.53} & \textcolor{red}{0.7895} & \textcolor{red}{27.78} & \textcolor{red}{0.7446} & \textcolor{red}{26.68} & \textcolor{red}{0.8057} & \textcolor{red}{31.17} & \textcolor{red}{0.9176} \\
        \hline
    \end{tabular}
    \vspace{2pt}
    \caption{PSNR and SSIM comparison with the state-of-the-art on five datasets. Best, second best , and third best performance are presented in \textcolor{red}{red}, \textcolor{blue}{blue}, and \textcolor{green}{green}.}
    \label{tab:quantitative_results}
\end{table}
\vspace{-1cm}
\subsection{Comparisons with the SOTA}
To validate the superiority of ML-CrAIST, we compare it against recent state-of-the-art methods (SOATs) under a scale factor of 2, 3, and 4, respectively. In particular, former works, VDSR \cite{kim2016accurate}, MemNet \cite{tai2017memnet}, EDSR \cite{lim2017enhanced}, SRMDNF \cite{zhang2018learning}, CARN \cite{ahn2018fast}, IMDN \cite{hui2019lightweight}, RFDN-L \cite{liu2020residual}, LatticeNet~\cite{luo2020latticenet}, SwinIR~\cite{liang2021swinir}, ESRT~\cite{lu2022transformer}, NGSwin~\cite{choi2023n}, and OmniSR~\cite{wang2023omni} are introduced for comparison.
\begin{itemize}
    \item[$\blacksquare$] \textbf{Quantitative results.} The quantitative results are presented in Table \ref{tab:quantitative_results}. In order to be fair comparison throughout the evaluation process, all models undergo training and testing processes using the same dataset. It is clear from the results that our method achieves the highest performance across all testing datasets. Compared to \cite{wang2023omni}, ML-CrAIST has 0.20 dB improvement on Manga109 ($\times 3$). Also, we noticed that our method demonstrates the most significant improvement on B100, Urban100, and Manga109 datasets compared to existing methods, indicating its suitability for images rich in textured regions, geometric structures, and finer details of SR images. As shown in Table \ref{tab:lpips}, we obtain a lower LPIPS score, suggesting a higher perceptual quality of the SR image. It is worth noting that by incorporating the frequency details and analyzing the features in multiple scales, ML-CrAIST surpasses the performance of the existing methods. Additionally, in Table \ref{tab:quantitative_results}, we have shown the results of our lighter method (\textbf{Ours-Li}) with reduced parameters and FLOPs. It takes the minimum FLOPs among all the existing schemes with comparable results. The FLOPs are $\sim 1.5\times$ lesser than NGSwin with $1.01\%$ and $0.37\%$ PSNR and SSIM gain on Manga109 ($4\times$). Further, We have shown the computational overhead during inference in Table \ref{tab:runtime}.
\begin{table}[ht]
    \scriptsize
    \centering
    \begin{tabular}{c|c|c|c|c|c}
        \hline
        Model & \textbf{Set5} & \textbf{Set14} & \textbf{B100} & \textbf{Urban100} & \textbf{Manga109} \\
        \hline
        IMDN & 0.1317 $\pm$ 0.0659 & 0.1242 $\pm$ 0.0866 & 0.1907 $\pm$ 0.0601 & 0.0131 $\pm$ 0.0124 & 0.0038 $\pm$ 0.0032 \\
        SwinIR & \textcolor{red}{0.1287 $\pm$ 0.0642} & 0.1209 $\pm$ 0.0870 & 0.1857 $\pm$ 0.0596 & 0.0111 $\pm$ 0.0106 & 0.0033 $\pm$ 0.0026\\
        NGSwin & 0.1291 $\pm$ 0.0640 & 0.1210 $\pm$ 0.0869 & 0.1861 $\pm$ 0.0595 & 0.0109 $\pm$ 0.0101 & 0.0035 $\pm$ 0.0029\\
        OmniSR & 0.1293 $\pm$ 0.0641 & 0.1193 $\pm$ 0.0848 & 0.1829 $\pm$ 0.0595 & 0.0102 $\pm$ 0.0093 & 0.0034 $\pm$ 0.0029\\
        \hline
        \textbf{Ours-Li} & 0.1354 $\pm$ 0.0651 & \textcolor{black}{0.1197 $\pm$ 0.0859} & \textcolor{black}{0.1842 $\pm$ 0.0595} & \textcolor{black}{0.0105 $\pm$ 0.0097} & \textcolor{black}{0.0033 $\pm$ 0.0028}\\
        \textbf{Ours} & 0.1312 $\pm$ 0.0642 & \textcolor{red}{0.1173 $\pm$ 0.0845} & \textcolor{red}{0.1812 $\pm$ 0.0591} & \textcolor{red}{0.0101 $\pm$ 0.0094} & \textcolor{red}{0.0032 $\pm$ 0.0027}\\
        \hline
        \end{tabular}
    \vspace{2pt}
    \caption{LPIPS score Comparison on $4\times$. Best performance is presented in \textcolor{red}{red}. Lower score is better.}
    \label{tab:lpips}
\end{table}
\vspace{-15pt}
\begin{table}[ht]
    \scriptsize
    \centering
    \begin{tabular}{c|c|c|c|c|c}
    \hline
    & & & \multicolumn{3}{c}{Inference time (second)}\\
    \hline
        scale & Input dimension & Output dimension& OmniSR & Ours-Li & Ours \\
        \hline
        $2\times$ & (512, 382) & (1024, 764) & 4.98 & \textbf{4.24} & 5.99 \\
        $3\times$ & (341, 254) & (1023, 762) & 2.73 & \textbf{2.12} & 3.35 \\
        $4\times$ & (256, 191) & (1024, 764) & 2.05 & \textbf{1.94} & 2.61 \\
        \hline
    \end{tabular}
    \caption{Single image inference time for $2\times$, $3\times$, and $4\times$, respectively}
    \label{tab:runtime}
\end{table}
\vspace{-10pt}
\begin{figure*}[ht]
    \centering
    \setlength\abovecaptionskip{-0.1\baselineskip}
    \includegraphics[width=0.9\textwidth, height = 9.5cm]{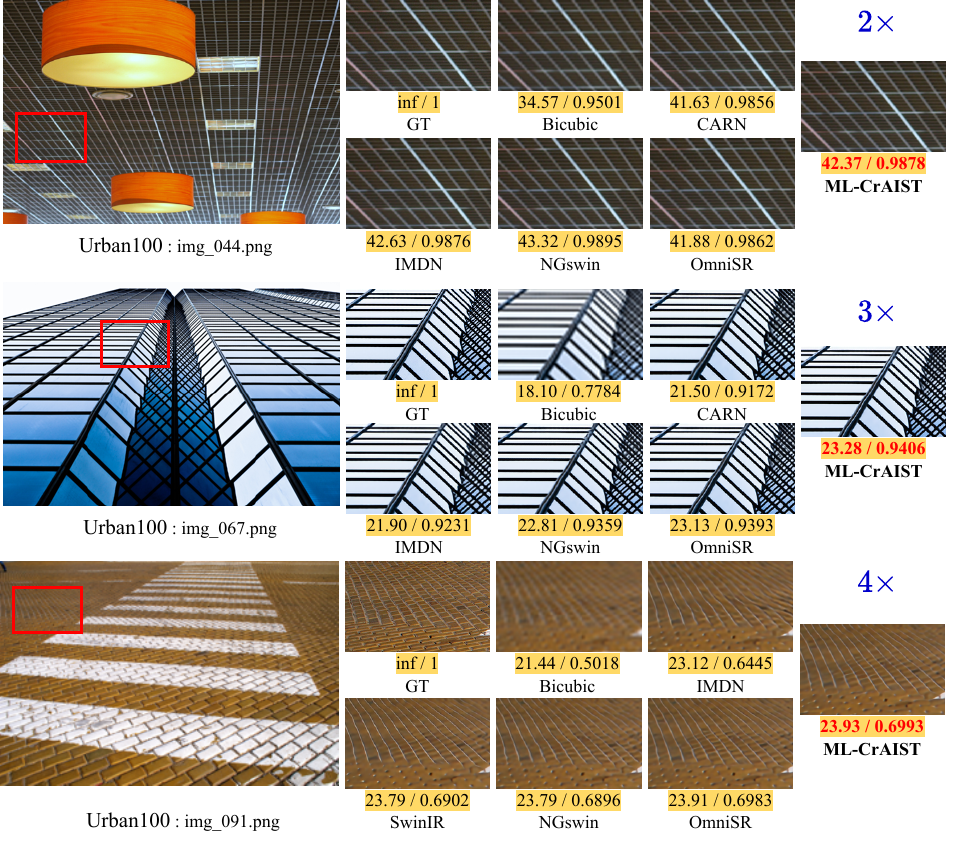}
    \caption{Visual Comparison of our ML-CrAIST with the SOTA.}
    \label{fig:vc1}
\end{figure*}
    \item[$\blacksquare$] \textbf{Visual Comparison.}
\begin{figure*}[ht]
    \centering
    \setlength\abovecaptionskip{-0.1\baselineskip}
    \includegraphics[width=0.7\textwidth, height = 2.6cm]{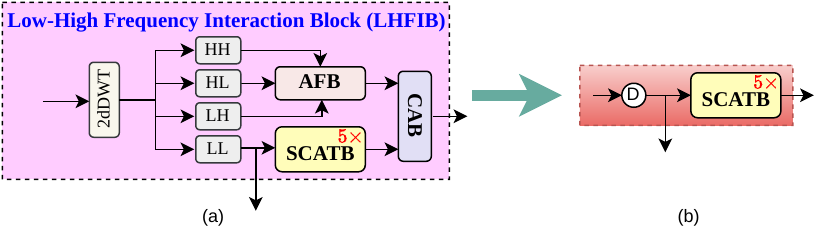}
    \caption{(a) indicates the LHFIB, (b) indicates the diagram without frequency information. \textcircled{\tiny D} indicates the bi-cubic down-sampling operation.}
    \label{fig:ab1}
\end{figure*}
\begin{figure*}[ht]
    \centering
    \setlength\abovecaptionskip{-0.1\baselineskip}
    \includegraphics[width=\textwidth]{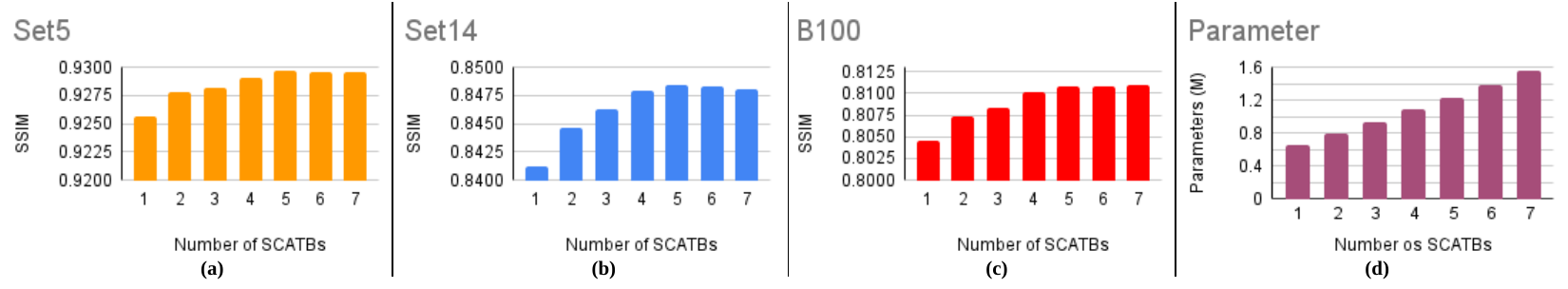}
    \caption{(a), (b), and (c) refer the SSIM comparison, and (d) refers the number of parameters on $3\times$ with different number of SCATBs.}
    \label{fig:ntm}
\end{figure*}
Figure \ref{fig:vc1} shows the visual comparison of our method with SOTAs at $\times 2$, $\times 3$, and $\times 4$ scales. It is observable that the HR images generated by ML-CrAIST exhibit more fine-grained details, whereas other methods produce blurred edges or artifacts in complex regions. For example, in the third image of Figure \ref{fig:vc1}, our model can pleasantly restore the precise texture of the road. The visual results demonstrate that incorporating frequency information and analyzing features across multiple scales enables us to capture more structural information, preserve the geometric structure of the image, and generate realistic HR results.
\end{itemize}
\vspace{-10pt}
\subsection{Ablation Study}
\begin{figure*}[!ht]
    \centering
    \setlength\abovecaptionskip{-0.1\baselineskip}
    \includegraphics[width=\textwidth]{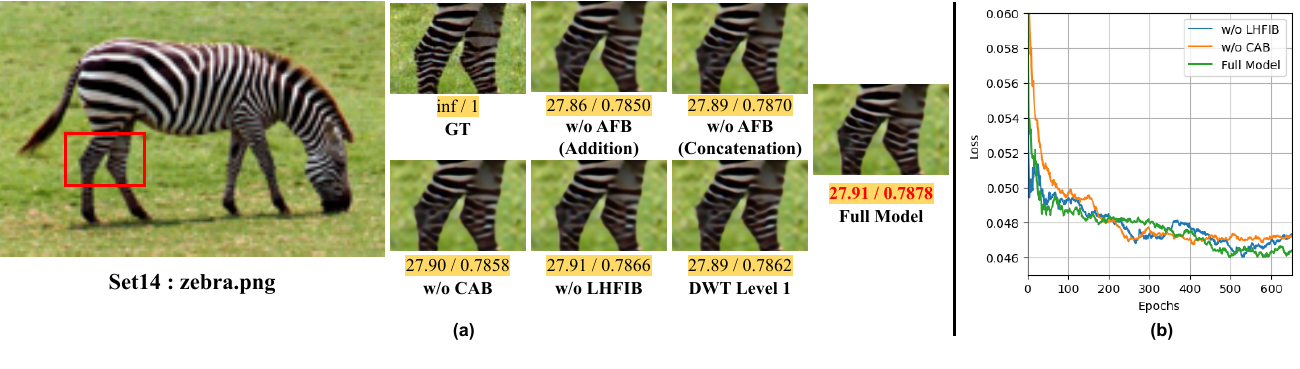}
    \caption{(a) Visual comparison of different settings of ML-CrAIST. (b) Convergence graph of ML-CrAIST.}
    \label{fig:vc2}
\end{figure*}

\begin{figure*}[!ht]
    \centering
    \setlength\abovecaptionskip{-0.1\baselineskip}
    \includegraphics[width=\textwidth]{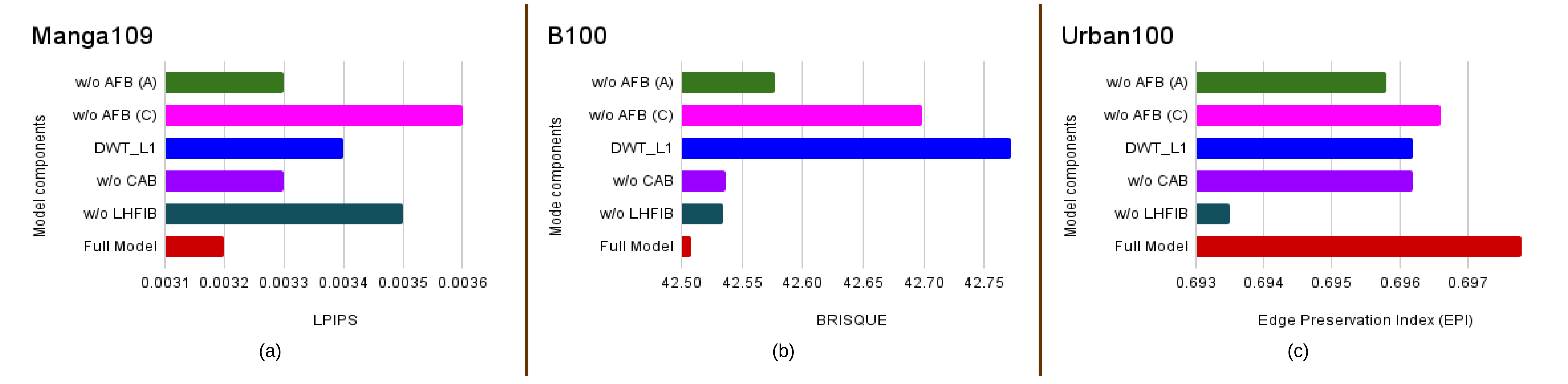}
    \caption{LPIPS ($\downarrow$), BRISQUE ($\downarrow$), and EPI comparison bettween different components of ML-CrAIST. $\downarrow$ indicates lower is better.}
    \label{fig:ebe}
\end{figure*}

In this subsection, we perform a set of experimentations to exhibit the efficacy of ML-CrAIST in different settings.
\begin{itemize}
    \item[$\blacksquare$] \textbf{Number of SCATBs.} Experimentally, we have set the number of SCATBs to 5. We also analyze the model performance by varying the SCATB number N. As depicted in Figure \ref{fig:ntm}, compared to the smallest number of SCATB, increasing the number of SCATB leads to performance gains. It can be seen that ML-CrAIST with $N = 6$ or $7$ produces a similar kind of result as $N = 5$ with higher parameters (refer to \ref{fig:ntm}(d)).
    \item[$\blacksquare$] \textbf{Effect of LHFIB.} We remove the frequency information and only take the spatial information to train our model. Figure \ref{fig:ab1}(a) and \ref{fig:ab1}(b) represent the diagram with and without the frequency information, respectively. The results are reported in the ${5}^{th}$ row of the Table \ref{tab:ab}. The results of ML-CrAIST are superior with the frequency information, displaying that the frequency details can offer global dependency to enhance the representation capability of the model. 

    \begin{table}[!ht]
    \scriptsize
    \centering
    \tiny
    \begin{tabular}{c|c|c|c|c|c|c|c|c|c|c|c}
        \hline
         & FLOPs &\multicolumn{2}{|c|}{\textbf{Set5}} & \multicolumn{2}{|c|}{\textbf{Set14}} & \multicolumn{2}{|c|}{\textbf{B100}} & \multicolumn{2}{|c|}{\textbf{Urban100}} & \multicolumn{2}{|c}{\textbf{Manga109}} \\
        \hline
        Model& (G) & PSNR & SSIM & PSNR & SSIM & PSNR & SSIM & PSNR & SSIM & PSNR & SSIM  \\
        \hline
        w/o AFB (Addition)& 42.80 & 32.28 & 0.8974 & \textcolor{black}{28.47} & 0.7886 & 27.56 & 0.7431 & 26.64 & 0.8041 & \textcolor{black}{31.14} & 0.9174\\
        w/o AFB (Concatenation)& 42.82 & 32.29 & 0.8974 & 28.45 & 0.7885 & 27.68 & 0.7435 & 26.63 & 0.8056 & 31.09 & 0.9174\\
        DWT Level-1 & 41.11 & 32.15 & 0.8957 & 28.46 & 0.7872 & 27.72 & 0.7423 & 26.58 & 0.8022 & 31.04 & 0.9157\\
        w/o CAB & 41.79& 32.31 & 0.8977 & 28.52 & 0.7881 & 27.76 & 0.7433 & 26.65 & 0.8043 & 31.10 & \textcolor{black}{0.9175}\\
        w/o LHFIB & 42.53& 32.29 & 0.8975 & 28.42 & 0.7888 & 27.26 & 0.7434 & 26.66 & 0.8050 & 31.11 & 0.9164 \\ 
        \hline
        \textbf{Full Model}& 42.91 &  \textcolor{red}{32.36} & \textcolor{red}{0.8984} & \textcolor{red}{28.53} & \textcolor{red}{0.7895} & \textcolor{red}{27.78} & \textcolor{red}{0.7446} & \textcolor{red}{26.68} & \textcolor{red}{0.8057} & \textcolor{red}{31.17} & \textcolor{red}{0.9176} \\
        \hline
        \end{tabular}
    \vspace{2pt}
    \caption{Ablation studies with different settings of our model on $4\times$. Best result is represented in \textcolor{red}{red}.}
    \label{tab:ab}
\end{table}
    \item[$\blacksquare$] \textbf{Effect of CAB.} We execute experiments to investigate the significance of the CAB. Specifically, we compare the results of the model with and without CAB in the $4^{th}$ row of Table \ref{tab:ab}. While removing the CAB, we used a simple element-wise addition operation.  From the aspects of quantitative metrics, the use of CAB can obviously improve the SSIM and PSNR performance of the model. The visual comparison is shown in Figure \ref{fig:vc2}(a). 

    \item[$\blacksquare$] \textbf{Effect of AFB.} We explore the feature aggregation process in the $1^{st}$ and $2^{nd}$ row of Table \ref{tab:ab}. The results demonstrate that the proposed AFB produces promising outcomes compared to summation and concatenation methods. 

    \item[$\blacksquare$] \textbf{Effect of multi-scale or multi-level DWT.} Third row of Table \ref{tab:ab} justifies the importance of the 2-level 2dDWT or multi-scale analysis in our model. 

    % \item[$\blacksquare$] \textbf{Runtime and Memory Usages.} The qualitative and quantitative results of ML-CrAIST have already been demonstrated in the above section. Now, we show the computational overhead during inference in Table \ref{tab:runtime}. Our model has $1259K/3.453G$, $1268K/3.504G$, and $1280K/3.576G$ parameters/FLOPs for $2\times$, $3\times$, and $4\times$, respectively.
\end{itemize}
Further, to validate each component of ML-CrAIST, in Figure \ref{fig:ebe}, we have shown results in three different measurements: LPIPS, Blind/Referenceless Image Spatial Quality Evaluator (BRISQUE), and Edge Preservation Index (EPI). It can be seen that the full model has a lower LPIPS and BRISQUE and a high EPI value, which indicates that the image has fewer distortions, artifacts, and better edge preservation, aligns more closely with natural scene statistics, and is visually pleasing to human observers. 
% \begin{figure}
%   \begin{minipage}[b]{.39\linewidth}
%     \centering
%     \includegraphics[width=\linewidth, height= 3.5cm]{SSIM vs FLOPs.pdf}
%     \caption{ FLOPs and SSIM comparison between our lighter ML-CrAIST model and SOTAs on Manga109 for 4x SR.}% \caption{Figure caption}
%   \end{minipage}\hfill
%   \begin{minipage}[b]{.6\linewidth}
%     \centering
%     \includegraphics[width=\linewidth]{edge_key.pdf}
%     \caption{Key-point and canny edge detection comparison between existing methods and ML-CrAIST. The top corner of the first row indicates the number of key points.}
%     \label{fig:edge}
%   \end{minipage}
% \end{figure}
\begin{figure*}[!ht]
    \centering
    \setlength\abovecaptionskip{-0.1\baselineskip}
    \includegraphics[width=0.8\textwidth, height = 3.5cm]{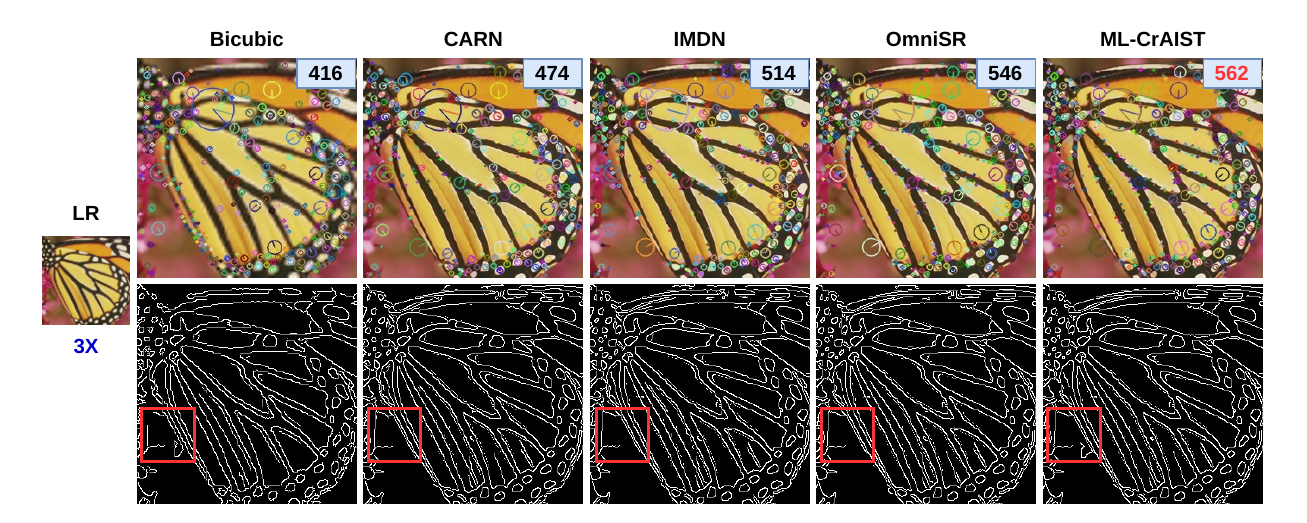}
    \caption{Key-point and canny edge detection comparison between existing methods and ML-CrAIST. The top corner of the first row indicates the number of key points.}
    \label{fig:edge}
\end{figure*}
\subsection{Impact on various application}
To validate the practical applicability of our model, we employ ML-CrAIST as a prepossessing technique for image key-point detection and edge detection tasks, as shown in Figure \ref{fig:edge}. Initially, we employ scale-invariant feature transform (SIFT) to compute the key points. It can be observed that the key-point detection significantly increases after super-resolving the images using our method. Subsequently, we employ Canny edge detection to identify edges in the super-resolved images. Compared to the super-resolved image by SOTA models, our super-resolved image exhibits more localized edge features. In the second row of Figure \ref{fig:edge}, we have marked using a red box where our method captures edges perfectly, but others fail. 
\section{Conclusion}
In this paper, we propose a transformer-based multi-scale super-resolution architecture called ML-CrAIST, demonstrating the advantage of modeling both spatial and frequency details for the SR task. Our cross-attention block seamlessly performs message passing between low and high-frequency features across multiple scales in the network and acknowledges their correlation. Furthermore, we propose AFB to effectively fuse the high frequency cubes, which boosts the overall performance. Finally, we validate the rationale and efficiency of the ML-CrAIST by conducting extensive experimentation across various benchmark datasets. We additionally conduct an ablation study to assess the impact of various configurations within ML-CrAIST.
\bibliographystyle{splncs04}
\bibliography{samplepaper}
\end{document}